\documentclass{llncs}
\usepackage[pdftex]{graphicx}
\usepackage[english]{babel}

\usepackage[ruled,vlined,noresetcount]{algorithm2e}

\usepackage{lscape}
\usepackage{framed}
\usepackage{url}
\usepackage{amsmath}
\usepackage{color, colortbl}
\usepackage[table]{xcolor}
  \usepackage{soul, color}
  \sethlcolor{green}

\usepackage{multirow}

 \usepackage[neveradjust]{paralist}

\begin{document}

\title{Toward verbalizing ontologies in isiZulu}
\author{C. Maria Keet$^1$ and Langa Khumalo$^{2}$}
\institute{$^1$Department of Computer Science, University of Cape Town, South Africa, \email{mkeet@cs.uct.ac.za}\\
$^2$Linguistics Program, School of Arts, University of KwaZulu-Natal, South Africa, \email{Khumalol@ukzn.ac.za}}
\maketitle

\begin{abstract}
IsiZulu is one of the eleven official languages of South Africa and roughly half the population can speak it. It is the first (home) language for over 10 million people in South Africa. Only a few computational resources exist for isiZulu and its related Nguni languages, yet the imperative for tool development exists. We focus on natural language generation, and the grammar options and preferences in particular, which will inform verbalization of knowledge representation languages and could contribute to machine translation. The verbalization pattern specification shows that the grammar rules are elaborate and there are several options of which one may have preference. We devised  verbalization patterns for subsumption, basic disjointness, existential and universal quantification, and conjunction. This was evaluated in a survey among linguists and non-linguists. Some differences between linguists and non-linguists can be observed, with the former much more in agreement, and preferences depend on the overall structure of the sentence, such as singular for subsumption and plural in other cases.
\end{abstract}

\section{Introduction}

While South Africa has been celebrated as having the most enabling constitution in the protection and advancement of African languages and has had a stable democracy for two decades, there is a glaring limitation in the investment in computational linguistics and human language technologies (HLT).  Although the imperative in HLT exist, there has been a lack of coordination in the development of HLT in African languages and a total lack of commitment from government and related institutions to invest and advance HLTs. 
However, the need for them is voiced; e.g., the ``National Recordal System'' project by the National Indigenous Knowledge Systems Office (NIKSO) of the South African Department of Science and Technology, and the University of KwaZulu-Natal, which recently made a ground-breaking introduction of mandatory isiZulu module for all its students, and is driving the development of scientific terminology in isiZulu. Visible advances are those that have been made by large companies such as Google and Microsoft, which have seen the localization of the user interfaces of their software, and an error-full Google Translate English-isiZulu. These and related endeavours require natural language generation and machine translation, and multilingualism in knowledge representation (e.g., \cite{Fogwill11,AFK12}) with end-user and domain expert interfaces, which do not exist. 

Multilingual systems are being developed elsewhere (among many, \cite{Bosca14,Kaljurand13,don06a}), and there are larger projects, such as Monnet ({\small \url{http://www.monnet-project.eu}}) for foundational aspects and Organic.Lingua ({\small \url{http://www.organic-lingua.eu}}) as applied project. As it appeared during our investigation, these advances are not immediately applicable with Nguni languages. Starting from the base and defining a grammar alike described in \cite{Kuhn13}, is a rather daunting task, because the still important reference for linguistic work for isiZulu and Southern Bantu languages are old and outdated \cite{Doke27,Doke35} and it will take many years to update. In the meantime, it is prudent to commence with the basics of NLG in such a way to serve linguists, computer scientists, and domain experts to show relevance. To this end, we take common language constructs of a practical logic language, such as the OWL 2 EL profile \cite{OWL2profiles} that is also used for the SNOMED CT medical terminology, as a starting point to focus on CNL and verbalizations of logical theories. 
Concerning verbalizations for OWL ontologies, it is already known that there are variations for verbalizations within English \cite{Schwitter08} and good English-OWL systems exist, notably ACE \cite{Fuchs10} and SWAT \cite{Third11}. The main results for logic-based conceptual models have been obtained also for monolingual English verbalizations \cite{Curland07}, whereas limitations of the so-called template-based approach have been investigated for the multilingual setting \cite{JKD06}. 

Overall, this raises several questions: 1) what are the verbalization patterns for isiZulu for the basic constructs? 2) what does that entail for an implementation of a verbalization? 3) are there theoretical options one can choose from, like in other languages, and which ones are preferred among isiZulu speakers? 
We devised the high-level patterns for verbalization of subsumption, disjointness, existential and universal quantification, and conjunction. The grammar rules for isiZulu are complex to the extent that a template-based approach is not feasible for either of the constructs investigated.  
This is due to, mainly, the semantics of the noun that affects several other components in a sentence, and the highly agglutinative nature of isiZulu. This also means existing multilingual and verbalization models and infrastructures cannot be transposed and translated to the Nguni languages. There are verbalization options, and we have conducted a survey to elicit both linguist and non-linguist preferences to inform algorithm development for a NLG-focussed grammar engine. 

The remainder of this paper is organised as follows. Section~\ref{sec:zuluIntro} describes some basic aspects of isiZulu and Section~\ref{sec:main} summarizes the  results of the verbalization patterns. The user evaluation is presented in Section~\ref{sec:eval}. We discuss in Section~\ref{sec:disc} and conclude in Section~\ref{sec:concl}.

\section{Some very basic aspects of isiZulu}
\label{sec:zuluIntro}


IsiZulu is the most populous language in South Africa spoken as a first (home) language by about 23\% of the over 50 million population. It has been documented for over a hundred and seventy years with the first booklet {\em Incwadi Yokuqala Yabafundayo} having been produced in 1837. It is a Bantu language that belongs to the Nguni sub-group of languages comprising of isiXhosa, isiNdebele, siSwati and isiZulu. Bantu languages have characteristically agglutinating morphology, which makes them rich and complex in their structure and one of the salient features of isiZulu is the system of noun classes. Each noun in isiZulu belongs to a noun class. It is the noun class that controls the concordance of all words in a sentence whose structure is typically subject verb object (SVO), although variations are attested to exist.  The isiZulu noun class prefixes, based on Meinhoff (1948), are mostly coupled in terms of singular/plural, and the classes are listed in Table~\ref{tab:nouns}. 

\begin{figure}[h]
\centering
   \includegraphics[width=0.45\textwidth]{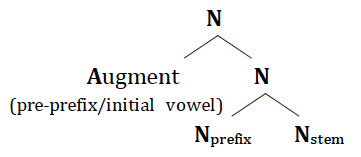} 
    \caption{The structure of isiZulu Nouns.}
    \label{fig:classtree}
\end{figure}

The noun comprises of two formatives, the prefix and the stem; its structure is depicted in Fig.~\ref{fig:classtree}. Prefixes express number and indicate the class to which a particular noun belongs. A prefix can be characterized as full or incomplete. The full prefix has the augment (pre-prefix) followed by a prefix proper and an incomplete prefix only has the augment respectively illustrated as {\em i}$_{\mbox{\small augment}}$ {\em mi}$_{\mbox{\small prefix proper}}$ {\em fula}$_{\mbox{\small stem}}$ ({\em imifula} = rivers) and {\em o}$_{\mbox{\small augment}}$ {\em gogo}$_{\mbox{\small stem}}$ ({\em ogogo} = grandmothers). Because of the agglutinating nature of isiZulu, a number of prefixes are phonologically conditioned and yet others are homographs. As stated earlier, the morphology of the head noun in the subject position will then influence the agreement pattern as shown in following the example.
\begin{description}
\item Amantombazana amadala adlala ibhola elimhlophe
\item {\bf ama}-ntombazana {\bf ama}-dala  {\bf a}-dlal-a  \hspace{8mm} {\bf i-}bhola \hspace{1mm}  {\bf e-li}-mhlophe
\item {\bf 6.}-girls  \hspace{15mm}   {\bf 6.}big  \hspace{6mm}         {\bf 6.SUBJ}-play {\bf 5.}-ball   {\bf REL-5}.-white
\item `The big girls are playing with the white ball'
\end{description}
The abbreviations by convention refer, respectively, to SUBJ = subject, REL = relative and 6./5. = noun class 6 and 5 respectively. 
The complex agreement system presents interesting challenges in the development of computational technologies in isiZulu. The understanding of the basic morphological structure of isiZulu is crucial in the formulation of the technologies.


\begin{table}[t]
\caption{Zulu noun classes, with examples. {\small The noun class prefix of classes 1 and 3 is conditioned by the morphology of the stem to which it attaches: {\em -mu-} before monosyllabic stems and {\em -m-} for other stems. 
C: Noun class, AU: augment, PRE: prefix, NEG SC: negative subject concord, PRON: pronomial.}}
\begin{center}
\begin{tabular}{|p{0.6cm}|p{0.7cm}|p{1cm}|p{1.5cm}|p{0.8cm}|p{1.1cm}|p{2.7cm}|p{1.7cm}|p{1.1cm}|}
\hline
{\bf C} & {\bf AU} & {\bf PRE} & {\bf Stem} (example) & {\bf NEG SC} & {\bf PRON} & {\bf Meaning} & \multicolumn{2}{|c|}{\bf Example} \\ \hline   \hline           
1 & u- & m(u)- & -fana  & aka- & yena & humans and other   & umfana & boy \\
 2 & a- & ba- & -fana & aba- & bona & animates & abafana & boys \\ \hline
  1a & u- & - & -baba & aka- & yena & kinship terms and & ubaba & father \\
   2a & o- & - & -baba & aba- & bona & proper names & obaba & fathers \\ \hline
     3a & u- & - & -shizi & aka- & wona & nonhuman & ushizi & cheese \\ 
      (2a) & o- & - & -shizi & aba- & bona &  & oshizi & cheeses \\ \hline
     3 & u- & m(u)- & -fula & awu- & wona & trees, plants, non- & umfula & river \\
     4 & i- & mi- & -fula & ayi- & yona & paired body parts & imifula & rivers \\ \hline
      5 & i- & (li)- & -gama & ali- & lona & fruits, paired body & igama & name \\
       6 & a- & ma- & -gama & awa- & wona & parts, natural phenomena  & amagama & names \\ \hline
        7 & i- & si- & -hlalo & asi- & sona & inanimates and  & isihlalo & chair \\
         8 & i- & zi- & -hlalo & azi- & zona & manner/style & izihlalo & chairs \\ \hline
         9a & i- & - & -rabha & ayi- & yona & nonhuman & irabha & rubber \\ 
          (6) & a- & ma- & -rabha & awa- & wona &  & amarabha & rubbers \\ \hline                  
          9 & i(n)- & - & -ja & ayi- & yona & animals & inja & dog \\
           10 & i- & zi(n)- & -ja & azi- & zona & & izinja & dogs \\ \hline
            11 & u- & (lu)- & -thi & alu- & lona & inanimates and & uthi & stick \\
             (10) & i- & zi(n)- & -thi & azi- & zona & long thin objects & izinthi & sticks \\ \hline
              14 & u- & bu- & -hle & abu- & bona & abstract nouns & ubuhle & beauty \\ 
               15 & u- & ku- & -cula & aku- & khona & infinitives & ukucula & to sing \\ \hline
               17 &  & ku- &  & & & locatives, remote/ general & & locative \\  \hline              
\end{tabular}
\end{center}
\label{tab:nouns}
\end{table}%

\vspace{-3ex}

\section{Summary of the relevant grammar rules}
\label{sec:main}

Due to space limitations, we describe the patterns and important features, not the analysis we have conducted and (elaborate) algorithms developed.

The essence of possible verbalizations of the quantifiers and the main connectives 
are shown in Table~\ref{tab:constructors}. The enumerations in the isiZulu column indicate that its use {\em depends on the context}, which may be the category or noun class it applies to, or other aspects in the axiom before or after the symbol, and the additional ``/'' within an item refers to the fact that one of them has to be chosen, depending on the noun class or first letter of a term. The main variables that affect verbalization in isiZulu for the cases we investigated are the noun class of the name of the OWL class and category of the OWL class, whether the OWL class is an atomic class or a class expression, the quantifier used in the axiom, and the position of the OWL class in the axiom. 

\begin{table}[!h]
\caption{A few constructors, their typical verbalization in English, and the basic options in isiZulu; see text for further details.}
\begin{center}
\begin{tabular}{|p{1.5cm}|p{2.4cm}|p{7.9cm}|}
\hline
{\bf DL symbol} & {\bf Sample verb. English} & {\bf Sample verbalization in isiZulu} \newline (see text for additional rules) \\ \hline \hline
$\sqsubseteq$ & ... is a ... & Depends on what is on the rhs of $\sqsubseteq$ and desideratum: \newline 
	\mbox{ } \hspace{1mm} A) semantic distinction\newline
		\mbox{ } \hspace{5mm} i) yi/ongu/uyi/ngu \hfill (living thing) \newline
		\mbox{ } \hspace{5mm} ii) iyi \hfill (non-living thing) \newline 	
	\mbox{ } \hspace{1mm} B) syntactic distinction\newline		
		\mbox{ } \hspace{5mm} iii) ng \hfill (nouns commencing with a, o, or u) \newline 		
		\mbox{ } \hspace{5mm} iv) y \hfill (nouns commencing with i) \\ \hline		
$\equiv$  & 1) ... is the same as ... \newline 2) ... is equivalent to ... 
            &  I. Depends on what is on the rhs of $\equiv$: \newline
		\mbox{ } \hspace{1mm} i) ufana no/ne \hfill (person) \newline
		\mbox{ } \hspace{1mm} ii) ifana/lifana/afana \hfill (not a person) \newline
		II. Depends on grammatical number on lhs of $\equiv$: \newline
		\mbox{ } \hspace{1mm} ii) yinto efanayo \hfill (singular) \newline
		\mbox{ } \hspace{1mm} ii) zifana ne/no/nezi \hfill (plural)
            \\ \hline
$\sqcup$   & ... or ... & 1) ... okanye ... \newline
					 2) ... noma ... \\ \hline
$\sqcap$    & ... and ... & Depends on the use of the $\sqcap$: \newline 
			\mbox{ } \hspace{1mm} i) ... na/ne/no ... \hfill (list of things) \newline 
			\mbox{ } \hspace{1mm} ii) 1) ... futhi ... \hfill  (connective) \newline
			\mbox{ } \hspace{5mm}  2) ... kanye ... \hfill  (connective) \\ \hline
$\neg$     & not ... & angi/akusiso/akusona/akubona/akulona/asibona/ akalona/akuyona \\ \hline
$\exists$       & 1) some ... \newline 2) there exists ... \newline 3) at least one ... & Depends on position in axiom:\newline
			I. quantified over class, depends on meaning of class: \newline
			\mbox{ } \hspace{1mm} i) kuno \hfill (living thing) \newline
			\mbox{ } \hspace{1mm} ii) 	kune \hfill (non-living thing) \newline
			II. includes relation (preposition issue omitted):\newline
			\mbox{ } \hspace{1mm} 1) ... [concords]dwa\newline
			\mbox{ } \hspace{1mm} 2) ... noma [copulative + concord]phi ... \newline							\mbox{ } \hspace{1mm} 3) thize 	
							 \\ \hline
$\forall$       & 1) for all ... \newline 2) each ... & Depends on what it is quantified over: \newline
		\mbox{ } \hspace{1mm} A) semantic distinction\newline
			\mbox{ } \hspace{5mm} i) wonke/bonke/sonke/zonke \hfill (living thing) \newline
			\mbox{ } \hspace{5mm} ii) onke/konke/lonke/yonke \hfill  (non-living thing) \newline
		\mbox{ } \hspace{1mm} B) another semantic distinction\newline
			\mbox{ } \hspace{5mm} i) use noun class  \hfill  (see Table~\ref{tab:nouns})			
							 \\ \hline
\end{tabular}
\end{center}
\label{tab:constructors}
\end{table}%

\vspace{-2ex}

\subsubsection{Subsumption.} There are two different ways of carving up the nouns to determine which rules apply (Table~\ref{tab:constructors}), but for generating good verbalizations, the main issue is to choose between singular or plural with or without the universal quantification voiced, which are illustrated in (S1). One can construct a similar set of options for generic (S2) versus determinate that has an extra {\em u-} (S3), but the generic is preferred for the neutral setting of verbalizations.
\vspace{-2ex}
\begin{description}
\item[(S1)] ${\tt\small MedicinalHerb \sqsubseteq Plant}$
\item {\sf ikhambi \underline{ng}umuthi} \hfill (`medicinal herb \underline{is a} plant')
\item {\sf amakhambi \underline{yi}mithi} \hfill (`medicinal herbs \underline{are} plants')
\item {\sf \underline{wonke} amakhambi \underline{ng}umuthi} \hfill (`\underline{all} medicinal herbs \underline{are a} plant')
\item[(S2)] ${\tt\small Giraffes \sqsubseteq Animals}$ 
\item {\sf izindlulamithi \underline{yi}zilwane} \hfill (`giraffes \underline{are} animals'; generic)
\item[(S3)] ${\tt\small Cellphone \sqsubseteq Phone}$
\item {\sf Umakhalekhukhwini \underline{uyi}foni} \hfill (`cellphone \underline{is a} phone'; determinate)
\end{description}
The possible patterns for subsumption can be, with $N$=noun taken from the name of the OWL class, and NC=noun class: 
\begin{compactenum}[a.]
\item $N_1$ $<$copulative {\em ng}/{\em y} depending on first letter of $N_2$$>$$N_2$.
\item $<$plural of $N_1$$>$ $<$copulative {\em ng}/{\em y} depending on first letter of plural of $N_2$$>$$<$plural of $N_2$$>$. 
\item $<$All-concord for NC$_x$$>$onke $<$plural of $N_1$, being of NC$_x$$>$ $<$copulative {\em ng}/{\em y}  depending on first letter of $N_2$$>$$N_2$.
\end{compactenum}
%
If the subsumption is followed by negation, then the copulative is omitted, and also here there are options; e.g.:
\vspace{-1ex}
\begin{description}
\item[(SN1)] ${\tt\small Cup \sqsubseteq \neg Glass}$
\item {\sf indebe \underline{akuyona} ingilazi} \hfill (`cup \underline{not a} glass')
\item {\sf \underline{zonke} izindebe \underline{aziyona} ingilazi} \hfill (`\underline{all} cups \underline{not a} glass')
\end{description}
\vspace{-1ex}
It combines the negative subject concord (NEG SC) of the noun class of the first noun ({\em aku-}) with the pronomial (PRON) of the noun class of second noun ({\em -yona}), where each noun class has its version (see Table~\ref{tab:nouns}). Thus, the pattern for negation in subsumption can be: 
\begin{compactenum}[a.]
\item  $<$$N_1$ of NC$_x$$>$ $<$NEG SC of NC$_x$$>$$<$PRON of NC$_y$$>$ $<$$N_2$ of NC$_y$$>$.
\item $<$All-concord for NC$_x$$>$onke $<$plural $N_1$, being of NC$_x$$>$ $<$NEG SC of NC$_x$$>$$<$PRON of NC$_y$$>$ $<$$N_2$ with NC$_y$$>$.
\end{compactenum}
%
%
We leave the more complicated cases where the inclusion axiom can be used, like $\forall R.C \sqsubseteq \exists S.(D \sqcap E)$, for future work, as well as negation in other contexts. 

\subsubsection{Conjunction.}
The `and' as enumeration uses {\em na}, which changes into (a + i =) {\em ne} or (a + u =) {\em no}, depending on the first letter of the noun and is then prefixed to the second noun that drops its first letter (always a vowel), illustrated in (C1). Conjunction as connective of clauses has two options, being  {\em kanye} or {\em futhi}, illustrated in (C2).  
\begin{description}
\item[(C1)] ${\tt\small Butter \sqcap Milk}$
\item {\sf Ibhotela \underline{no}bisi} \hfill ({\em Ibhotela} + {\em na} + {\em Ubisi})
%
\item[(C2)] ${\tt\small \ldots \exists has\_filling.Cream \sqcap \exists has\_Icing.Lemon\_flavour \ldots}$
\item {\sf ...kune zigcwalisa ukhilimu \underline{kanye} nezinye uqweqwe olunambitheka\_ulamula...}
\item {\sf ...kune zigcwalisa ukhilimu \underline{futhi} nezinye uqweqwe olunambitheka\_ulamula...}
\end{description}
While this distinction is a hassle in the algorithm, the pattern is straightforward.

\subsubsection{Existential Quantification.} Option I in Table~\ref{tab:constructors} refers to cases like (E0), which are not used in OWL ontologies, but that has axioms of type (E1) instead, which include the object property (verb) and for which there are several verbalization options. 
\begin{description}
\item[(E0)]  {\sf Ezulwini \underline{kune} zingilosi} \hfill(`in heaven \underline{there exist} angels')
\item[(E1)] ${\sf Giraffe \sqsubseteq \exists eats.Twig}$
\item {\sf izindlulamithi zindla izihlamvana} \hfill (`giraffes eat twigs')
\item {\sf yonke indlulamithi idla ihlamvana \underline{elilodwa}} \hfill (`each giraffe eats \underline{at least one} twig')
\item {\sf zonke izindlulamithi zidla ihlamvana \underline{elilodwa}} \hfill (`all giraffes eat \underline{at least one} twig')
\item  {\sf yonke indlulamithi idla  \underline{noma yiliphi} ihlamvana} \hfill (`each giraffe eats \underline{some} twig')
\item  {\sf zonke izindlulamithi zidla  \underline{noma yiliphi} ihlamvana} \hfill (`all giraffes eat \underline{some} twig')
\item  {\sf yonke indlulamithi idla ihlamvana\underline{thize}} \hfill (`each giraffe eats \underline{some} twig')
\end{description}
Thus, we have a choice between `at least' and `some', and singular and plural.
The quantification (underlined text) is more important than verb conjugation here. For the `at least one',  the relative concord (RC) is determined by the noun class system and is attached to the quantitative concord (QC) and then suffixed with the quantitative suffix {\em -dwa}; e.g.:\\
%
\vspace{-5mm}
\begin{table}[!h]
\begin{center}
\begin{tabular}{llcccccccc} 
noun && NC && RC & QC & QSuffix & copulative & EP & ESuffix\\ \hline
{\em ihlamvana} (`twig') && class 5 && {\em eli-} & {\em -lo-} & {\em -dwa} & & &\\
{\em isifundo} (`module') && class 7&& {\em esi-} & {\em -so-} & {\em -dwa} & & & \\
{\em ushizi} (`cheese') && class 3a && {\em o-} & {\em -ye-} & {\em -dwa} & & & \\
{\em ihlamvana} (`twig') && class 5 && &  & & {\em yi-} & {\em -li-} & {\em -phi}\\
{\em isifundo} (`module') && class 7 && &  & & {\em yi-} & {\em -si-} & {\em -phi}\\
{\em ushizi} (`cheese') && class 3a && &  & & {\em ngu-} & {\em -mu-} & {\em -phi}
\end{tabular}
\end{center}
\end{table}
\vspace{-5mm}
%
%
 
\noindent There are lookup tables for that, like for the NEG SC and PRON in Table~\ref{tab:nouns}. 
For the `some' option, it is constructed as copulative + enumerative prefix (EP) + enumerative suffix {\em -phi}, as illustrated above, and the conjunction {\em noma} collocates with the enumerative to complete the meaning `some among many'. Also for the EP, there is a fixed concord for each noun class. Note that the {\em -i}, respectively {\em -u}, are added to the copulative, because the copulative cannot be followed by a consonant that the EP begins with. 
Finally, the clitic {\em -thize}, which has a variant form {\em -thile}, is another expression of the complex morphology. The clitic {\em -thize} attaches to the noun, which is often the object of the sentence, to express the sense that it is some among many of those objects. In (E1) {\em inhlamvanathize} would thus mean any one of the twigs. This is a bit borderline in meaning, but it is the only candidate for being a template for that aspect. Overall, the following three core patterns are obtained:
\begin{compactenum}[a.]
\item $<$All-concord for NC$_x$$>$onke $<$pl. $N_1$, is in NC$_x$$>$ $<$conjugated verb$>$ $<$$N_2$ of NC$_y$$>$ $<$RC for NC$_y$$>$$<$QC for NC$_y$$>$dwa;
\item  $<$All-concord for NC$_x$$>$onke $<$pl. $N_1$, is in NC$_x$$>$ $<$conjugated verb$>$ noma $<$copulative {\em ng/y} adjusted to first letter of $N_2$$>$$<$EP of NC$_y$$>$phi $<$$N_2$$>$.
\item $<$All-concord for NC$_x$$>$onke $<$$N_1$ in NC$_x$$>$ $<$conjugated verb$>$ $<$$N_2$$>$thize;
\end{compactenum}
Verb conjugation is a separate matter, which is complicated to encode, but there are no options to choose from in a verbalization. This is also the case for the prepositions in an OWL object property name (like the `by' in `taught by').

\section{Experimental evaluation of the verbalisation patterns}
\label{sec:eval}

The aim of the experiment is to show how the understanding of the basic structure of isiZulu can illuminate the verbalization in isiZulu. While there are various options to verbalize something in isiZulu and these options involve elaborate algorithms,  the experiment sought to find out possible preferences for the verbalization of the subsumption, disjointness, and quantifiers in isiZulu from the participants. The experiment also sought to find out if variations in verbalizations mattered to different participants, in particular between linguists and non-linguists. 

\subsection{Survey design}

The set up of the experiment was as follows.
\begin{compactenum}[1)]
\item Devise a set of sentences that tests the patterns introduced in Section~\ref{sec:main}, include a few cross-checks, add an `either' and `neither' option, and add auxiliary question, being whether the participant is a linguist or not, optional comments and optional contact email. The sentences will be generated through manual application of the patterns. For instance, Question 1 asks the participant to choose between: 
\begin{compactenum}[a)]
\item {\em Ikhambi ngumuthi}; subsumption singular
\item {\em Amakhambi yimithi}; subsumption plural
\item {\em Wonke amakhambi ngumuthi}; with `for all' quantifier, and plural
\item {\em Yomithathu}; either one of them
\item {\em Awukho}; neither
\end{compactenum}
Question 2 asks the same thing as in Question 1, but then with the giraffes. Question 3 offers the option for disjointness singular versus plural and the universal quantification verbalised. Question 4: also asks about disjointness: 
\begin{compactenum}[a)]
\item {\em Ihebhivo alilona ikhanivo}; singular (disjointness of herbivore and carnivore)
\item {\em Amahebhivo awalona ikhanivo}; plural
\item {\em Yomibili}; either one of the two
\item {\em Awukho}; neither
\end{compactenum}
Questions 5 to 9 deal with existential quantification: {\em -dwa} versus {\em noma ...-phi} (Question 5); Question 6 fixes {\em noma ...-phi} but varies by singular vs plural; Question 7 does the same but then with  {\em -dwa}:
\begin{compactenum}[a)]
\item {\em Sonke isifundo sifundiswa nguSolwazi oyedwa}; singular (each course is taught by at least one professor)
\item {\em Zonke izifundo zifundiswa nguSolwazi oyedwa}; plural
\item for ``either'' and d) ``neither''
\end{compactenum} 
Question 8 pits them against each other with singular {\em -dwa} vs singular {\em noma ...-phi} vs plural {\em noma ...-phi}, and Question 9 requires a choice on plural {\em noma ...-phi} versus {\em -thize}. Question 10 asks about {\em kanye} vs {\em futhi}. The complete list of sentences is included in the supplementary file, and the survey is left open for people to consider, accessible at {\small \url{http://limesurvey.cs.ukzn.ac.za/index.php?sid=25965&lang=zu}} (click the button labelled `okulandelayo' to proceed to the main set of questions).
\item Set up the survey in the locally installed Limesurvey ({\small \url{http://www.limesurvey.org}}). This was chosen because we had localised the relevant part of it to isiZulu in a previous research activity and no other survey software has a localization to isiZulu. Thus, all questions, answer options, autotext, and error messages are in isiZulu. 
\item Invite people via email to participate. 
\item Collect data after 2.5 days, and analyse it using MS Excel. 
\end{compactenum}

\subsection{Results and discussion}

Twenty five people were invited to participate in the survey, among them students, academics (linguists), and non-linguists, such as administrators. In the short time frame, this resulted in 12 completed responses, 5 of whom self-identified as linguist and 7 as non-linguist. The results are depicted in Table~\ref{tab:survey} and the (anonymised) excel file is accessible at {\small \url{http://www.meteck.org/files/CNLsurvey.xls}}, which also has a copy of the questions and the full answer options; here, question ``1. isa'' is Question 1 as described above in the materials and methods, and ``sing.'' corresponds to the first option {\em Ikhambi ngumuthi}, ``pl.'' to {\em Amakhambi yimithi} and so on, and likewise for Question 4's ``sing.'' being the {\em Ihebhivo alilona ikhanivo} option listed above as answer option a). 

The survey results show clear preferences from linguists. Question 1 option 1 is an overwhelming choice and this is predictable because of its simple structure. Once the nominal head takes the plural form there seems to be hesitation because of the complex agreement system; e.g., the answers given to Question 2 illustrates this. In the context of negation (Questions 3 and 4), the answers were mixed overall, though there was a slight preference among non-linguists for simple singular over simple plural (not present with the nominal head, Question 3). This agreement system was unavoidable for the `forall-exists' pattern, and there the plural is preferred over singular (Questions 7 and 8). The {\em -phi} and {\em -thize} (Question 9) seem not to have any preferences, while {\em -dwa} has overwhelming preference over {\em -phi} (Question 5). 

It is an attested fact that there are dialects of isiZulu and some of the salient differences in preferences may be based on dialect variation, which also may explain the dislike for either {\em kanye} or {\em futhi} (Question 10). It is important that the survey remains open to more participants and maybe a clearer pattern of preferences may emerge. A further option could be to ask them whether the generated sentence is understandable (rather than preference), which may become more relevant when larger axioms are going to be verbalised.


\begin{table}[!h]
\caption{Survey results in percentage of votes (rounded) and disaggregated by linguist (Ling.) and non-linguist (Non-Ling.); sing.: singular; pl.: plural; all: $\forall$ verbalised; exists: with the `forall-exists' construction; and: the connective-and.}
\begin{center}
\begin{tabular}{|p{1.2cm}|p{2cm}|p{0.7cm}|p{0.7cm}|p{0.8cm}||p{1.2cm}|p{2cm}|p{0.7cm}|p{0.7cm}|p{0.8cm}|}
\hline
\multicolumn{2}{|c|}{\bf Question} & \multicolumn{3}{|c||}{\bf Respondent} & \multicolumn{2}{|c|}{\bf Question} & \multicolumn{3}{|c|}{\bf Respondent} \\ 
\multicolumn{2}{|c|}{\bf } &  Ling. & Non-Ling. & Total & \multicolumn{2}{|c|}{\bf } & Ling. & Non-Ling. & Total 
\\ \hline \hline
\multirow{5}{*}{1. isa}  & sing. & 80 & 0 & 33 & \multirow{5}{*}{6. exists} & sing.+noma-phi & 0 & 29 & 17 \\  \cline{2-5}\cline{7-10}
	& pl. & 0 & 43 &  25& & pl.+noma-phi & 0 & 0 & 0 \\ \cline{2-5}\cline{7-10}
	& all+pl. & 0  & 0 & 0 & & either & 20 & 0 & 8 \\ \cline{2-5}\cline{7-10}
	& either & 20 & 57 &  42& & neither & 80 & 71 & 75 \\ \cline{2-5}\cline{7-10}
	& neither & 0  & 0 & 0 & & & & & \\ \hline
\multirow{5}{*}{2. isa}  & sing. & 80 & 86 & 83 & \multirow{5}{*}{7. exists} & sing.+-dwa & 20 &14 & 17 \\  \cline{2-5}\cline{7-10}
	& pl. & 0 & 0 & 0 & & pl.+-dwa & 20 & 57 & 42 \\ \cline{2-5}\cline{7-10}
	& all+pl. & 0 & 0 & 0 & & either & 40 & 0 &17  \\ \cline{2-5}\cline{7-10}
	& either & 0 & 14 & 8 & & neither & 20 & 29 & 25 \\ \cline{2-5}\cline{7-10}
	& neither & 20 & 0 & 8 & & & & & \\ \hline
\multirow{5}{*}{3. disj.}  & sing. & 40 & 29 & 33 & \multirow{5}{*}{8. exists} & sing.+-dwa & 0 & 14 & 8 \\  \cline{2-5}\cline{7-10}
	& all+pl. & 0 & 14 & 8 & & sing.+noma-phi & 20 &0 & 8 \\ \cline{2-5}\cline{7-10}
	& either & 40 & 14 & 25 & & pl.+noma-phi & 80 & 57  & 67 \\ \cline{2-5}\cline{7-10}
	& neither & 20 & 43 & 33 & & either & 0 & 0 & 0 \\ \cline{2-5}\cline{7-10}
	& & & &  & & neither & 0 & 29 & 17 \\ \hline
\multirow{4}{*}{4. disj.}  & sing. & 40 & 71 & 58 & \multirow{4}{*}{9. exists} & pl.+noma-phi & 40 & 14 & 25 \\  \cline{2-5}\cline{7-10}
	& pl. & 0 & 0 & 0 & & pl.+-thize & 0 & 29 & 17 \\ \cline{2-5}\cline{7-10}
	& either & 20 & 0 & 8 & & either & 40 & 43 & 42 \\ \cline{2-5}\cline{7-10}
	& neither & 40 & 29 & 33 & & neither & 20 & 14 & 16 \\ \hline	
\multirow{4}{*}{5. exists}  & pl.+-dwa & 100 & 57 & 75 & \multirow{4}{*}{10. and} & kanye & 0 & 0 & 0 \\  \cline{2-5}\cline{7-10}
	& pl.+noma-phi & 0 & 14 & 8 & & futhi & 0 & 14 & 8 \\ \cline{2-5}\cline{7-10}
	& either & 0 & 0 & 0 & & either & 20 & 0 & 8 \\ \cline{2-5}\cline{7-10}
	& neither & 0 & 29 & 17 & & neither & 80 & 86 & 83 \\ \hline	
\end{tabular}
\end{center}
\label{tab:survey}
\end{table}%

\section{Discussion}
\label{sec:disc}

For languages with isolating morphology such as English, verbalization templates are known to be a good way to explore with developing a controlled natural language, or may even suffice for a use case scenario, whereas this approach breaks down for languages with richer (agglutinating) morphology \cite{JKD06}. So far, we have not found a single case where a plain template without supporting encodings of the grammar suffices.The survey revealed some general preferences, such as the {\em -dwa} option cf {\em -phi}, and dislikes ({\em futhi}/{\em kanye}), and a few notable differences in preference between linguists and non-linguists, such as the clear preference for the singular for subsumption and the unanimous preference for {\em -dwa} among the linguists, and overall the linguists agreed more on their preference compared to the non-linguists. The latter may have to do also with dialect and which region the isiZulu speaker is from. 

These results can already feed into the development of a verbalization tool, but it requires more research before committing to invest in something like the Grammatical Framework ({\small \url{http://www.grammaticalframework.org/}}). This paper highlights motivational use cases for further investigation that benefits both isiZulu linguistics and ICT and we will continue to extend the algorithms, add more, and implement them so that domain experts can contribute and easily access, among others, the indigenous knowledge management knowledge base and scientific terminologies in isiZulu.

As an added benefit, these results may also inform translation algorithms and tools. Take, e.g.,  Google Translate English-isiZulu machine translation online and enter `all giraffes eat some twig' is translated as {\em yonke izindlulamithi udle igatsha} (translation obtained 27-3-2014):  {\em izindlulamithi} is in noun class 10, so it should be {\em zonke} instead of {\em yonke}, and it misses the quantifier. This can be correctly verbalised with the pattern described in Section~\ref{sec:main}. As an aside, and looking toward additional verbalization patterns: its {\em udle} (2nd or 3rd person singular, imperative) is also incorrect, because the verb has to be conjugated for the noun class of the first noun, which is 3rd person plural non-human (in casu, {\em zidla}). 
%
Another aspect to investigate in detail is the living/non-living thing distinction, which we avoided for subsumption by using a syntax-based short-cut. This may, or may not, work for other verbalization patterns. 

Last, there are other issues to consider for verbalizing in isiZulu, such as multilingual ontologies and, possibly modifying the {\em Lemon} model \cite{McCrae12} to cater for annotation with, at least, the noun class, and finding ways for semi-automated ontology translations to/from isiZulu. 


\section{Conclusions}
\label{sec:concl}

Verbalizing formally represented knowledge in isiZulu requires a grammar engine even for the relatively basic language constructs, which are due to, mainly, the noun classes, the agglutinative nature of isiZulu, and contextual knowledge about the position of the symbol in the axiom. The salient features peculiar to isiZulu that pose a challenge are: i) the system of noun classes, ii) the system of complex agreement, iii) phonological conditioned copulatives, and iv) verb conjugation. We developed a set of possible verbalization patterns for simple subsumption, disjoint classes, quantification, and conjunction. The survey on verbalization pattern preference revealed a clear preference for the {\em -dwa} option, and more variation in preference by the non-linguists.   

Algorithms have been developed for the verbalizations \cite{KeetK14rule}, which will be extended further for other larger axioms and verb conjugation, and implemented. It will also be useful to investigate comprehension of the generated sentences and the effect of dialect on preferences.


\subsubsection{Acknowledgements.} We thank N. Hadebe, Y. Motloung, M. Ndaba, and S. Nkosi for their initial exploration into the topic.


\end{document}